(REVIEW ARTICLE)

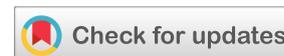

# Beyond the model: Key differentiators in large language models and multi-agent services

Muskaan Goyal [1, *] and Pranav Bhasin [2]

[1] Department of Computer Science, University of California, Berkeley, United States.
[2] Department of Electrical Engineering, University of California, Berkeley, United States.

## Abstract

With the launch of foundation models like DeepSeek, Manus AI, and Llama 4, it has become evident that large language models (LLMs) are no longer the sole defining factor in generative AI. As many now operate at comparable levels of capability, the real race is not about having the biggest model but optimizing the surrounding ecosystem, including data quality and management, computational efficiency, latency, and evaluation frameworks. This review article delves into these critical differentiators that ensure modern AI services are efficient and profitable.

## 1. Introduction

The rapid evolution of generative AI has led to a saturation point where numerous industry and open-source Large Language Models (LLMs) exhibit similar quality levels [1, 2]. If LLMs are no longer the competitive edge, what drives the advantage in AI services? The actual value in generative AI lies not in the models themselves but in the ancillary components that enhance and support these models.

Agentic tools like Cohere's Cline, an IDE plugin that links language models to company-specific data and adds relevant context to responses during coding or content creation, highlight the growing focus on the systems that support and enhance language model outputs. Cline demonstrates how important ancillary components—such as data retrieval, indexing, and context integration—are in improving the reliability and relevance of these models. This shift marks a move from developing new models to building the tools and frameworks that make them more practical, accurate, and adaptable for real-world enterprise applications.

## 2. Real Differentiators

### 2.1. Data Quality and Proprietary Datasets

Companies leading in AI aren't always the ones with the biggest models, but rather those with access to high-quality, domain-specific data. The effectiveness of AI systems comes from their ability to use this targeted data for better outcomes. Organizations with proprietary data can fine-tune models to meet their unique needs, giving them a significant competitive advantage [3, 4].

* Corresponding author: Muskaan Goyal

Model Reliability can be enhanced by integrating human-generated data through human-in-the-loop approaches to reach desired outcomes. Techniques like Retrieval-Augmented Generation (RAG) help reduce AI hallucinations and lower computational costs by dynamically retrieving information, enabling AI to provide more accurate responses without frequent retraining [5].

## 2.2. Efficiency: Computational Efficiency and Cost Optimization

Early foundation models indicated a significant demand for GPUs to train large language models, translating to high costs. For example, training GPT-3 was estimated to cost around $4.6 million [6]. However, innovations like DeepSeek have challenged this notion. DeepSeek's models claim to use significantly less computing power than Llama 3.1, leading to substantial reductions in computing cost and energy consumption [7, 8].

Several optimization strategies have emerged to enhance computational efficiency:

### 2.2.1. Model Quantization

Reducing the precision of model weights (e.g., from 32-bit to 4-bit) can significantly decrease memory usage and increase inference speed with minimal accuracy loss [9].

### 2.2.2. Model Pruning

Removing redundant or less significant weights from the model can reduce its size and computational requirements without substantially impacting performance [10].

### 2.2.3. Neural Attention Memory Models (NAMMs)

Developed by Sakana AI, NAMMs optimize the memory usage of LLMs by deciding which tokens to retain or discard, reducing unnecessary information, and improving performance. Experiments with the Meta Llama 3-8B model showed that NAMMs improved performance on natural language and coding tasks while saving up to 75% of cache memory [11].

### 2.2.4. Semantic Caching

By reusing responses to semantically similar queries, semantic caching can avoid redundant model calls and significantly save inference costs. Implementing an efficient semantic cache can reduce API calls by up to 68.8% [12].

### 2.2.5. Attention Offloading

This technique offloads Large Language Models' attention computation to memory-optimized devices while utilizing high-end accelerators for other model parts. This heterogeneous setup helps enhance efficiency and cost-effectiveness [13].

## 2.3. Latency and Costs for Business Use Case

Latency and operational costs are key factors defining the profitability and usability of AI applications in real-world applications. Advancements in model efficiency contribute to reducing inference times and energy consumption while ensuring prompt responses.

Techniques like speculative decoding, where smaller models draft responses that larger models then verify, can reduce latency [14]. Additionally, parameter-efficient fine-tuning methods, such as Low-Rank Adaptation (LoRA), update only a subset of weights during training, drastically cutting memory requirements [15].

Furthermore, frameworks like Flash-LLM enable cost-effective and highly efficient generative model inference by leveraging unstructured sparsity, significantly reducing GPU memory consumption and computation while retaining model accuracy [16].

## 2.4. Evaluation Frameworks and Monitoring

With the proliferation of new models and multi-agent architectures, robust evaluation frameworks are critical for assessing model performance, prompt updates, and system updates. A comprehensive evaluation framework ensures reliability and facilitates continuous improvement as new models are released.

Some recent developments include Scale Evaluation (a tool used to test advanced AI models across various benchmarks and tasks) [17], AILuminate (a tool introduced by MLCommons to benchmark models based on their responses to 24,000 test prompts across categories like inciting violence, hate speech, self-harm, and IP infringement) [18], and FrugalGPT (a strategy that learns which combinations of LLMs to use for different queries to reduce cost and improve accuracy) [19].

Companies can implement comprehensive monitoring systems to detect anomalies and measure effectiveness, resulting in sustained high performance in dynamic environments.

**2.5. Data Management Strategies**

More and more engineers today are focusing their time on managing data workflows—whether it's cleaning, labeling, augmenting, or fine-tuning pipelines for both real and synthetic data. It is becoming more common for engineering effort to go into handling data than into building new model architectures.

Some of the emerging strategies in this space include:

*2.5.1. Model-to-Data Movement*

Instead of moving large volumes of data to centralized compute clusters, there's a growing shift toward deploying lightweight or specialized models directly where the data is. This approach reduces bandwidth usage, safeguards sensitive data (critical in regulated industries), and helps lower latency.

*2.5.2. Synthetic Data Generation*

Companies like Gretel AI and Mostly AI are leading the way in generating synthetic datasets that maintain statistical properties while eliminating personally identifiable information (PII) [20]. This technique ensures safer model training and improves diversity by covering more edge cases in the data [21].

*2.5.3. Data Versioning and Lineage*

Open-source tools such as Data Version Control (DVC), LakeFS, and Pachyderm are becoming essential for ensuring reproducibility and traceability in the data used for training. These tools are critical in regulated industries and ensure auditability [22, 23].

## 3. Conclusion

Generative AI is undergoing a paradigm shift from model-centric development to ecosystem-centric innovation. As LLMs become increasingly commoditized, the real differentiators will be shaped by how effectively organizations can manage data, optimize compute costs, reduce latency, evaluate reliability, and maintain intelligent pipelines.

The future of LLM-based services lies in lean, sustainable, and highly operable infrastructure that supports agility and experimentation. This infrastructure enables businesses to derive tangible value from AI without the heavy baggage of over-engineering. Organizations that master these ecosystem levers will lead the next wave of generative AI breakthroughs.

**Compliance with ethical standards**

*Disclosure of conflict of interest*

There are no conflicts of interest to be disclosed.